\documentclass{article}
\usepackage{spconf,amsmath,graphicx,hyperref,amsfonts}

\title{Toward Robust Early Detection of Alzheimer's Disease via an Integrated Multimodal Learning Approach}
\name{
\parbox{\linewidth}{\centering
Yifei Chen$^1$, Shenghao Zhu$^1$, Zhaojie Fang$^1$, Chang Liu$^1$, Binfeng Zou$^1$, Linwei Qiu$^2$, \\ \textit{Yuhe Wang}$^1$, \textit{Shuo Chang}$^1$, \textit{Fan Jia}$^1$, \textit{Feiwei Qin}$^{1,\star}$, \textit{Jin Fan}$^1$, \textit{Yong Peng}$^1$, \textit{Changmiao Wang}$^3$}
}
\address{$^1$Hangzhou Dianzi University, Hangzhou, China \\ $^2$Beihang University, Beijing, China  \\ $^3$Shenzhen Research Institute of Big Data, Shenzhen, China \\
\tt$\!\!\!\star$ Corresponding Author: qinfeiwei@hdu.edu.cn\\}

\begin{document}

\maketitle

\begin{abstract}
Alzheimer's Disease (AD) is a complex neurodegenerative disorder marked by memory loss, executive dysfunction, and personality changes. Early diagnosis is challenging due to subtle symptoms and varied presentations, often leading to misdiagnosis with traditional unimodal diagnostic methods due to their limited scope. This study introduces an advanced multimodal classification model that integrates clinical, cognitive, neuroimaging, and EEG data to enhance diagnostic accuracy. The model incorporates a feature tagger with a tabular data coding architecture and utilizes the TimesBlock module to capture intricate temporal patterns in Electroencephalograms (EEG) data. By employing Cross-modal Attention Aggregation module, the model effectively fuses Magnetic Resonance Imaging (MRI) spatial information with EEG temporal data, significantly improving the distinction between AD, Mild Cognitive Impairment, and Normal Cognition. Simultaneously, we have constructed the first AD classification dataset that includes three modalities: EEG, MRI, and tabular data. Our innovative approach aims to facilitate early diagnosis and intervention, potentially slowing the progression of AD. The source code and our private ADMC dataset are available at \href{https://github.com/JustlfC03/MSTNet}{https://github.com/JustlfC03/MSTNet}. 
\end{abstract}

\section{Introduction}
\label{sec:intro}
Alzheimer's Disease (AD) is an irreversible, progressive neurodegenerative disorder characterized by comprehensive dementia, including memory impairment, executive dysfunction, and changes in personality and behavior \cite{c20hermann2022rapidly}. This debilitating condition significantly impairs a patient's thinking, memory, and independence, severely impacting daily life. As the global population ages, the prevalence of AD is increasing, posing substantial challenges to the economy, healthcare system, patients, and their families. Diagnosing AD in clinical settings is complicated by the subtlety of early symptoms, the diversity of symptom presentations, lengthy detection processes, and inconsistent diagnostic criteria, which are often highly dependent on a physician's expertise. Mild Cognitive Impairment (MCI) represents a critical window period before the onset of dementia \cite{c21karr2018does}. During this stage, the brain undergoes minor and often unnoticed changes, leading to frequent misdiagnosis. Therefore, timely and accurate diagnosis for patients in this period has great significance.


Recently, deep neural network techniques have significantly advanced the classification of AD, overcoming the limitations of traditional manual diagnosis methods. Research in this field can be broadly divided into two categories: unimodal and multimodal classification tasks. Unimodal diagnosis often falls short in capturing the full complexity of AD, resulting in diagnostic challenges and high rates of misdiagnosis. Conversely, multimodal approaches integrate multiple types of data, including clinical information, cognitive scales, neuroimaging, and EEG signals, to provide a more comprehensive understanding of the disease. Cognitive scales, for example, offer a rapid and precise assessment of intellectual status and cognitive deficits \cite{c22frey2004scholastic}. Neuroimaging, particularly Magnetic Resonance Imaging (MRI), provides detailed images of brain structures, allowing for the observation of changes in brain volume and cortical atrophy. Electroencephalograms (EEG) objectively reflect brain neural activity, containing rich physiological and pathological information \cite{c23luhmann2022early}. The integration of these multimodal data types enhances the accuracy and robustness of models.

\section{Related work}
Several studies have demonstrated the effectiveness of multimodal classification approaches in diagnosing AD, but neither make full use of multimodal information. 
For instance, Liu et al. \cite{c3liu2023monte} introduced a Monte Carlo integrated neural network using ResNet50, achieving 90\% accuracy. Wang et al. \cite{c4wang2023hypergraph} utilized a multimodal learning framework with graph diffusion and hypergraph regularization, reaching 96.48\% accuracy by integrating phenotypic features with genetic data. Elazab et al. \cite{c24elazab2024alzheimer} reviewed AD diagnosis using machine and deep learning models, highlighting multimodal data fusion techniques and recent advancements. Subsequently, Liu et al. \cite{c6liu2023cascaded} developed a cascaded multimodal hybrid Transformer architecture, achieving an AUC of 0.994.Additionally, Lei et al. \cite{c7lei2024alzheimer} created the FIL-DMGNN model, which effectively mitigated modal competition, resulting in excellent diagnostic outcomes. Dwivedi et al. \cite{c8dwivedi2022multimodal} proposed a novel method for fusing MR and PET images, achieving high accuracy rates using a dual support vector machine. Lastly, Fang et al. \cite{c19fang2024gfe} introduced the GFE-Mamba model, which integrates MRI and PET data through a 3D GAN-ViT architecture, demonstrating high accuracy in predicting AD.

\section{Method}
\begin{figure*}[htb]
\centerline{\includegraphics[width=0.80\textwidth]{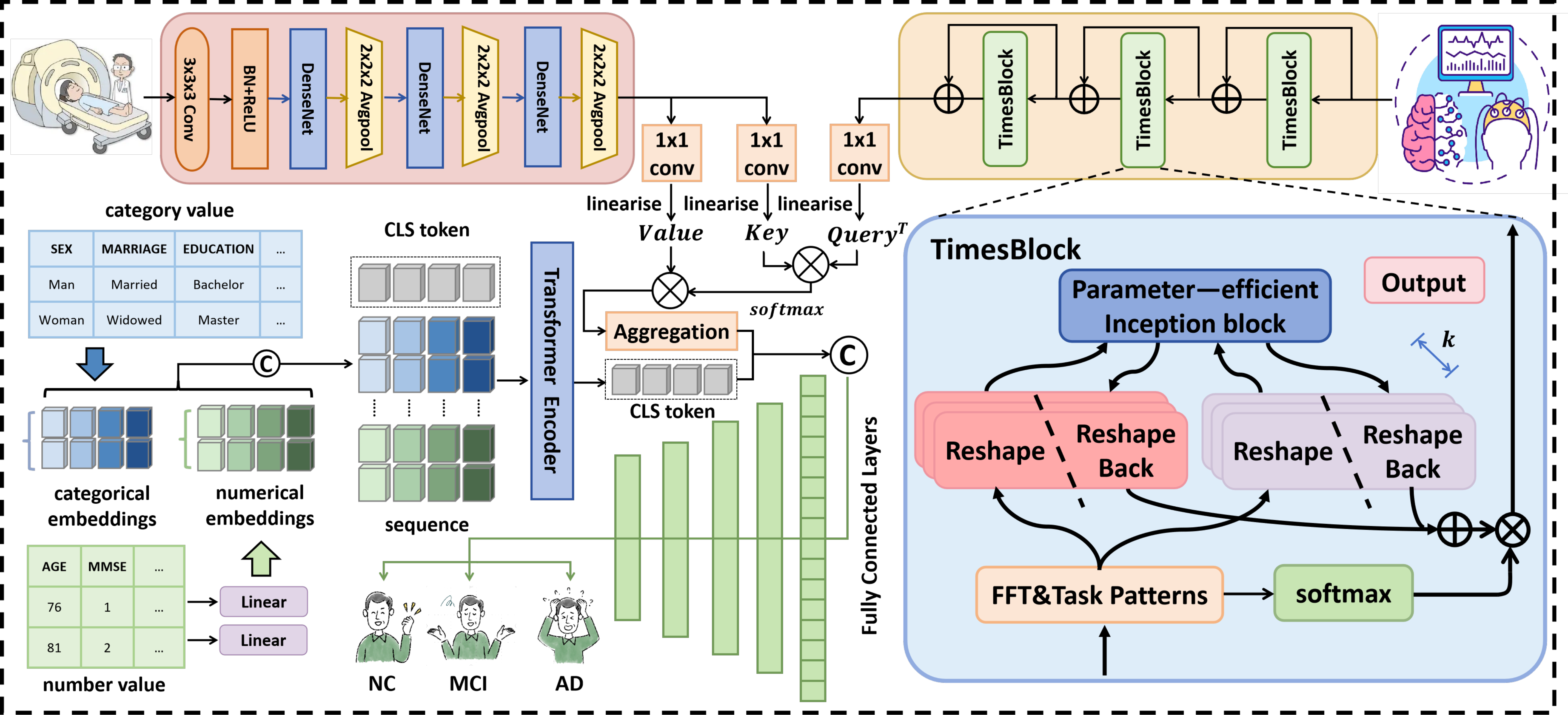}}
\vspace{-1.0em}
\caption{Overall framework of the proposed MSTNet. The MSTNet model primarily comprises three parts: Tabular Feature Encoder, Temporal Feature Encoder and Cross-modal Aggregation Encoder.}
\label{fig1}
\end{figure*}

As illustrated in Figure \ref{fig1}, we propose an enhanced multimodal classification model. This model integrates a feature tagger with a coding architecture designed specifically for tabular data. Furthermore, we employ the TimesBlock module to process EEG data, which effectively models complex temporal patterns by performing 2D transformations on the multi-periodic features of the temporal data. To further enhance performance, we utilize a Cross-modal Attention Aggregation module (CMAA). This module fuses the spatial information from MRI data with the temporal information from EEG data, capturing the intricate correlations.

\vspace{-1.0em}
\subsection{Tabular Feature Encoder}
In the diagnostic process of AD, tabular data such as patient clinical data and cognitive test results contain valuable information. However, the diverse nature of these data types poses challenges in processing them uniformly, often resulting in suboptimal diagnostic accuracy. Traditional methods have struggled to fully leverage the richness of this information. Our multimodal classification model addresses these challenges by efficiently processing tabular data using a feature tagger and Transformer architecture, inspired by FT-Transformer \cite{c11gorishniy2021revisiting}. By integrating the feature tagger with an encoding architecture tailored for tabular data, our model enhances the representation of specific modalities, thereby improving diagnostic accuracy.


\vspace{-1.0em}
\subsubsection{Feature Tokenizer}
As illustrated in Figure \ref{fig2} (a), the primary function of the feature tokenizer is to convert numerical features (such as a patient's age and MMSE score) and categorical features (such as gender and education) into embedding vectors. These embeddings are then input into the coding architecture designed for tabular data for feature extraction. The Feature Tokenizer transforms the input features \( x \) into embeddings \( T \in \mathbb{R}^{k \times d} \). The embedding for a given feature \( x_j \) is computed as follows:
\begin{equation}
    T_j = b_j+f_j(x_j)\in \mathbb{R}^d \hspace{10pt} f_j:\rightarrow \mathbb{R}^d,
\end{equation}
where \( b_j \) is the bias for the \( j \)-th feature, and this process is called Feature Biases. For numerical features, \( f_j^{(\text{num})} \) is implemented as an element-wise multiplication with the vector \( W^{(\text{num})} \in \mathbb{R}^d \), and for categorical features, \( f_j^{(\text{cat})} \) is implemented as a lookup table \( W_j^{(\text{cat})} \in \mathbb{R}^{S_j \times d} \) as follows:
\begin{small}
    \begin{equation}
        \begin{aligned}
&T_j^{(num)} = b_j^{(num)}+x_j^{(num)}\cdot W_j^{(num)} &\in \mathbb{R}^d,\\
&T_j^{(cat)} = b_j^{(cat)}+e_j^T W_j^{(cat)} &\in \mathbb{R}^d,\\
&T= \texttt{stack}[T_1^{(num)},\cdots ,T_{k^{(num)}}^{(num)},T_1^{(cat)}
,\cdots ,T_{k^{(cat)}}^{(cat)}] &\in \mathbb{R}^{k\times d},
        \end{aligned} 
    \end{equation}
\end{small}
where $e_j^{T}$ is a one-hot vector for the categorical feature.

\vspace{-1.0em}
\subsubsection{Tabular Encoder Architecture}
The Tabular Encoder Architecture processes the embedding vectors generated by the feature tokenizer. It captures complex relationships between features using a multi-head self-attention mechanism and a feed-forward network. The final output of the Tabular Encoder is derived from the processed output of the Transformer layer. In the Transformer's output, the [CLS] token summarizes the entire input sequence. This token is passed to the prediction layer, where it is combined with aggregated features from the image and EEG modalities via cross-modal attention. We also employ the PreNorm variant for easier optimization. In this setting, it is necessary to remove the first normalization from the first Transformer layer to achieve optimal performance.

\vspace{-1.0em}
\subsection{Temporal Feature Encoder}
\begin{figure*}[htb]
\centerline{\includegraphics[width=0.70\textwidth]{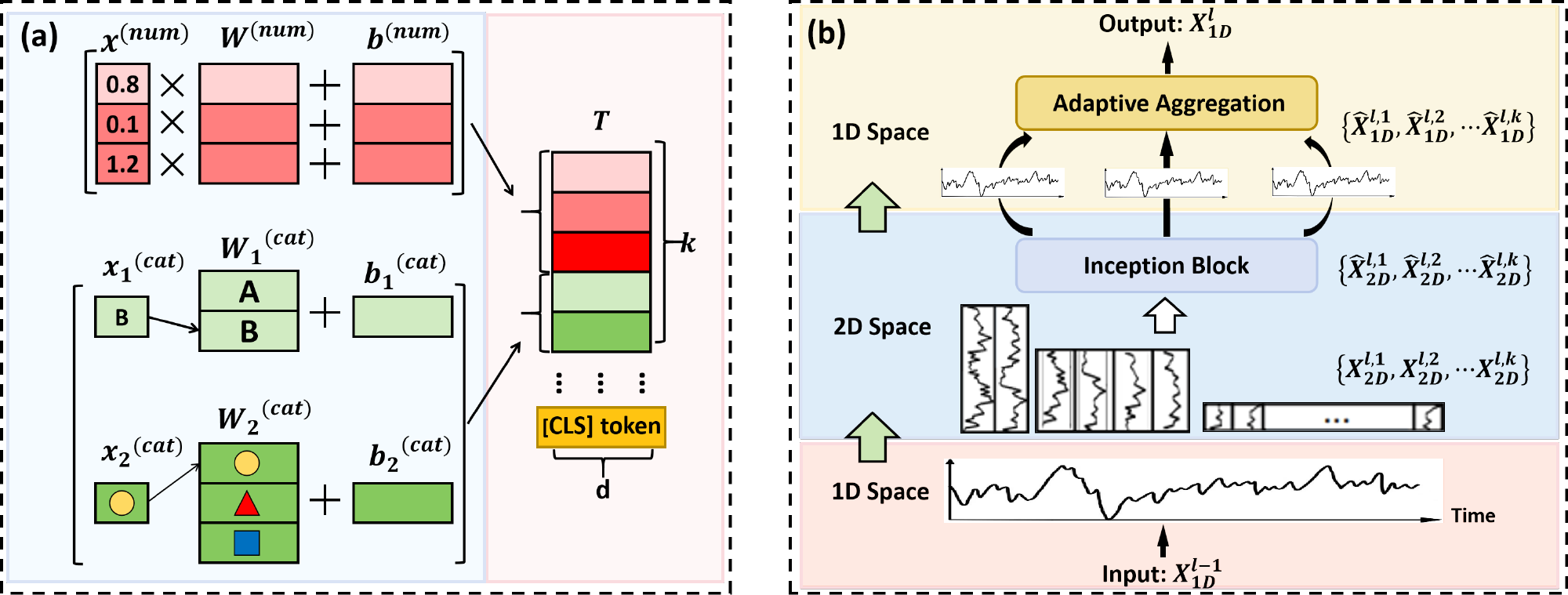}}
\vspace{-1.0em}
\caption{\textbf{(a)} The Feature Tokenizer converts numerical categorical features into embedding vectors. \textbf{(b)} The TimesBlock module performs a 2D transformation on the multi-periodic features of the time-series data.}
\label{fig2}
\end{figure*}

EEG data contain rich temporal information but are often plagued by noise and irrelevant details. Inspired by Wu et al. \cite{c12wu2022timesnet}, we incorporate the TimesBlock module to handle EEG data. As shown in Fig. \ref{fig2} (b), the TimesBlock module effectively models complex temporal patterns by performing a 2D transformation on the multi-periodic features of the time-series data, decomposing complex time-varying patterns into intra-periodic and inter-periodic variations.

The process begins by transforming a one-dimensional time series \( X_{1D} \in \mathbb{R}^{T \times C} \), where \( T \) is the time length and \( C \) is the channel dimension, using a fast Fourier transform (FFT) to calculate periodicity:
\begin{align}
A = \text{Avg}(\text{Amp}(\text{FFT}(X_{1D}))), \\
f_1, \cdots, f_k = \text{arg Topk}(A) \quad \substack{ f_* \in \{1, \cdots, [\frac{T}{2}]\} }, \\
p_1, \cdots, p_k = [\frac{T}{f_1}, \cdots, \frac{T}{f_k}],
\end{align}
where \( A \in \mathbb{R}^T \) represents the intensity of each resolved frequency component in \( X_{1D} \). The \( k \) frequencies \( \{f_1, \cdots, f_k \} \) with the largest intensity correspond to the most significant cycle lengths \( \{p_1, \cdots, p_k \} \). This process extracts key cycle features and filters out noise. Next, the original time series \( X_{1D} \) is folded based on the selected cycles:
\begin{equation}
X^i_{2D} = \text{Reshape}_{p_i, f_i}(\text{Padding}(X_{1D})), \quad i \in \{1, \cdots, k\},
\end{equation}
where \( \text{Padding}(\cdot) \) extends the time series by adding zeros along the time dimension. This ensures that the sequence length is divisible by \( p_i \). As a result, a set of two-dimensional tensors \( \{X_{2D}^1, X_{2D}^2, \cdots, X_{2D}^k\} \) is obtained, each corresponding to specific temporal variations. These 2D tensors exhibit locality, which can be captured by 2D convolutions. We use a multi-scale 2D convolution kernel to process the transformed 2D features and capture period variations (columns) and inter-period variations (rows) within the EEG signal:
\begin{equation}
\hat{X}_{l,i,2D} = \text{Inception}(X_{l,i,2D}),
\end{equation}
where the \( \text{Inception} \) block contains multiple 2D convolution kernels of different scales. This multi-scale convolution captures various feature patterns in the time series data, enhancing the model's ability to understand and process EEG signals. After processing, the 2D features are transformed back to 1D features with adapted and weighted according to the magnitude value of each cycle:
\begin{align}
\hat{X}_{l,i,1D} = \text{Trunc}(\text{Reshape}_{1, (p_i) \times f_i}(\hat{X}_{l,i,2D})),\\
X_{l,1D} = \sigma^k_{i=1} \hat{A}_{l-1, f_i} \times \hat{X}_{l,i,1D},
\end{align}
where \( \hat{A}_{l-1, f_i} \) is the magnitude value of the corresponding frequency, normalized by Softmax. This weighted aggregation ensures the significance of each periodic feature. To ensure robust training, we also apply residual concatenation to help mitigate gradient vanishing and explosion issues. Finally, features from multiple cycles are fused through dynamically adaptive aggregation of each periodic feature:
\begin{equation}
X_{l,1D} = \sigma^k_{i=1} \hat{A}_{l-1, f_i} \times \hat{X}_{l,i,1D}.
\end{equation}

\vspace{-1.0em}
\subsection{Cross-modal Attention Aggregation Module}
MRI data provide spatial information, while EEG signals offer temporal insights. Combining them allows for a more comprehensive analysis. We use a CMAA module to fuse MRI and EEG features effectively, capturing complex correlations in challenging multimodal tasks.

Initially, 3D MR images are processed using the DenseBlock module to capture detailed brain structural information, such as cortical atrophy and volume changes. Concurrently, EEG features extracted by the TimesNet Block are scaled into 3D features, enhancing temporal information representation. Cross-modal attention is then applied to the processed MRI and EEG features. The advanced features obtained through cross-modal attention are then combined with the simple concatenated features in the aggregation module. Finally, in the aggregation module, integrated features further capture deep relationships through residual dense connectivity. 




\begin{table*}
\vspace{-1.0em}
\caption{Comparison and ablation experiment of our proposed MSTNet on our private ADMC dataset.}
\label{tab:comparison and ablation}
\centering
\begin{tabular}{p{0.2\linewidth}p{0.12\linewidth}p{0.12\linewidth}p{0.12\linewidth}p{0.12\linewidth}p{0.12\linewidth}}
\hline
Method                                           & Precision & Recall  & F1-score      & Accuracy                            & MCC     \\ 
\hline
COMET \cite{c18wang2023contrast}  & 87.78  & 87.74 & 87.76 & 88.22 & 75.52 \\ 
JD-CNN \cite{c16abbas2023transformed} & 88.34  & 87.84 & 88.33 & 88.32 & 85.79 \\
TableTransformer \cite{c14huang2020tabtransformer}  & 88.10  & 88.73 & 87.78 & 89.10 & \textbf{87.52} \\
Zhang et al's \cite{c13zhang2023alzheimer} & 37.60  & 55.56  & 44.85   & 55.56 & 43.52   \\
Qiu et al's \cite{c5qiu2020development}  & 37.60  & 55.56   & 44.85  & 55.56 & 43.52   \\
Rad et al's \cite{c17article}. & 85.71  & 83.13  & 84.40 & 83.16 & 67.52 \\ \hline
w/o DenseBlock module  & 60.60  & 72.22  & 65.90  & 75.00 & 67.53 \\
w/o TimesBlock module  & 72.41  & 77.78  & 75.00    & 80.00                        & 74.35 \\
w/o CMAA module & 71.83  & 73.61  & 72.71 & 75.00                        & 63.57 \\
w/o Feature Biases      & 83.11  & 83.33  & 83.22 & 85.00 & 77.57 \\
\textbf{MSTNet (Our)} & \textbf{88.25}  & \textbf{88.89} & \textbf{88.57} & \textbf{90.00} & 86.16 \\ \hline
\end{tabular}
\vspace{-1.0em}
\end{table*}

\section{Experiment}
\vspace{-1.0em}
\subsection{Dataset and Experimental Details}
Our experiments were conducted on our private ADMC dataset, which comprises EEG, MRI, and scale data from 100 subjects. The subjects' demographic details are as follows: mean age of 72.4 years, age range from 56 to 93 years, with 56 females and 22 married individuals. The dataset is divided into 80 samples for training and 20 samples for evaluation. We intend to release this dataset publicly to support additional research in the field. We implemented the MSTNet network model using the PyTorch framework and trained it on an NVIDIA Tesla V100 GPU. During training, we set the dropout rate to 0.1, the batch size to 100, and the learning rate to 0.0001. The training process lasted for 24 hours.

\vspace{-1.0em}
\subsection{Comparison Experiment}
As presented in Table \ref{tab:comparison and ablation}, we compare the proposed MSTNet model with several advanced unimodal and multimodal models. The unimodal models include HTCF \cite{c18wang2023contrast}, JD-CNN \cite{c16abbas2023transformed}, and TableTransformer \cite{c14huang2020tabtransformer}. Among the multimodal models, we consider those developed by Zhang et al. \cite{c13zhang2023alzheimer}, Qiu et al. \cite{c5qiu2020development}, and Rad et al. \cite{c17article}. The MSTNet model, equipped with the CMAA module, demonstrates superior performance over the unimodal models in almost all metrics evaluated. Furthermore, MSTNet solves existing multimodal classification models, which often face challenges in multimodal feature extraction and fusion, demonstrating exceptional results.

\vspace{-1.0em}
\subsection{Ablation Study}
We conducted a series of ablation experiments to assess the impact of the DenseBlock module, TimesBlock module, CMAA module, and Feature Biases on the performance of MSTNet. The results are detailed in Table \ref{tab:comparison and ablation}. The experiments reveal that the removal of each module results in varying degrees of performance decline. Specifically, the absence of the DenseBlock module severely hampers the model's capability to process and extract intricate visual features from images. Similarly, removing the TimesBlock module significantly diminishes the model's ability to capture temporal features from the EEG signal. The exclusion of the CMAA module leads to a reduction in multimodal integration capabilities. Additionally, the model's stability is compromised without the Feature Biases. These findings underscore the essential role of each module in maintaining the overall effectiveness of our proposed advanced MSTNet model.

\section{Conclusion}



In this study, we introduce an advanced multimodal classification model designed for the diagnosis of AD. This model overcomes the limitations of traditional single-modality approaches by effectively integrating diverse data types. Our model has demonstrated superior performance in classifying AD, MCI, and NC, as confirmed by experiments conducted on our private ADMC dataset. This study reflects the potential of multimodal deep learning models in improving AD diagnosis and emphasizes the importance of integrating multiple data sources to gain a comprehensive understanding and prediction of the disease, representing a significant advancement in the field of neurodegenerative disease classification.

\vspace{-1.0em}
\section{Acknowledgment}
This work was supported in part by the National Key Research and Development Program of China (2023YFE0114900), the MoE Humanities and Social Sciences Project (24YJCZH225), Guangdong Basic and Applied Basic Research Foundation (No. 2022A1515110570), Guangxi Key R\&D Project (No. AB24010167), Innovation Teams of Youth Innovation in Science, Technology of High Education Institutions of Shandong Province (No.2021KJ088), Shenzhen Science and Technology Program (No. KCXFZ20201221173008022), and Shenzhen Stability Science Program 2022 (2023SC0073).

\vfill\pagebreak


\bibliographystyle{IEEEbib}

\end{document}